\documentclass[a4paper, 10pt, conference]{ieeeconf}

\newif\ifoagmfinalcopy

\oagmfinalcopytrue  
\IEEEoverridecommandlockouts                       

\usepackage{graphics} % for pdf, bitmapped graphics files
\usepackage{epsfig} % for postscript graphics files
\usepackage{mathptmx} % assumes new font selection scheme installed
\usepackage{times} % assumes new font selection scheme installed
\usepackage{amsmath} % assumes amsmath package installed
\usepackage{amssymb}  % assumes amsmath package installed
\usepackage{phonetic}
\usepackage{textgreek}
\usepackage{lineno}
\usepackage{tikzpagenodes}
\usepackage{background}
\usepackage{amsfonts}
\usepackage[mathscr]{euscript}

\ifoagmfinalcopy
\backgroundsetup{color=white}

\title{\LARGE \bf
Case Study: Ensemble Decision-Based Annotation of Unconstrained Real Estate Images
}

\ifoagmfinalcopy
\author{Miroslav Despotovic$^{1}$, Zedong Zhang$^{1}$, Eric Stumpe$^{2}$ and Matthias Zeppelzauer$^{2}$
\thanks{*This research was funded by the Austrian Research Promotion Agency
(FFG) project 880546 “IMREA” and we are very grateful to DataScience Service
GmbH for providing the data for this study.}
\thanks{$^{1}$M. Despotovic and Z. Zhang are with the Kufstein
University of Applied Sciences, Kufstein 6330, Tirol, Austria {\tt\small (miroslav.despotovic@fh-kufstein.ac.at;
zedong.zhang@fh-kufstein.ac.at)
}}%
\thanks{$^{2}$E. Stumpe and M. Zeppelzauer are with the ICMT Institute of Creative Media Technologies, St. Pölten University of Applied Sciences, 
St. Pölten 3100, Lower Austria, Austria
 {\tt\small 
(estumpe@fhstp.ac.at;
matthias.zeppelzauer@fhstp.ac.at)}}%
}

\else
\author{Anon, Ymous}
\fi

\begin{document}

\maketitle

%\ifoagmfinalcopy
%\thispagestyle{empty}
%\pagestyle{empty}
%\else
%\thispagestyle{plain}
%\pagestyle{plain}
%\begin{tikzpicture}[remember picture,overlay]
%\node[align=center,text=blue] at ([yshift=1em]current page text area.north) {\LARGE \#\#\# ARW/OAGM 2019 SUBMISSION: CONFIDENTIAL REVIEW COPY \#\#\#};
%\end{tikzpicture}%
%\fi

%%%%%%%%%%%%%%%%%%%%%%%%%%%%%%%%%%%%%%%%%%%%%%%%%%%%%%%%%%%%%%%%%%%%%%%%%%%%%%%%
\begin{abstract}

We describe a proof-of-concept for annotating real estate images using simple iterative rule-based semi-supervised learning. In this study, we have gained important insights into the content characteristics and uniqueness of individual image classes as well as essential requirements for a practical implementation.    

\end{abstract}

%%%%%%%%%%%%%%%%%%%%%%%%%%%%%%%%%%%%%%%%%%%%%%%%%%%%%%%%%%%%%%%%%%%%%%%%%%%%%%%%
\section{INTRODUCTION}
The annotation of unlabeled images is an important task for the assignment of metadata, which can be particularly challenging within a given knowledge domain.  
Thus, image metadata is being increasingly used in real estate research, e.g., for valuation \cite{Zhang2018}, location analysis \cite{Muhr2017}, or for estimating the condition of a building \cite{Koch2018}.
In the scientific literature, there are very few contributions on the classification of unlabeled images in the domain of real estate \cite{Poura2015}.
In this short paper, we present an approach to semi-supervised labeling of images containing interior and exterior views of real estate using simple ensemble classification rule.

\section{PROBLEM STATEMENT}

To maximize the information potential of the data, it must be tagged with meaningful labels, which in practice can require considerable manual effort.  
A typical approach for annotating unlabeled data autonomously is semi-supervised learning  (SSL), where an initial training set of labeled data $\emph{\hausaD}_\emph{\textiota}$ is defined by clustering and/or manual selection and the trained model is used to infer unlabeled data $\emph{\hausaD}_\emph{\textupsilon}$ systematically without interactively querying the user (e.g. active learning with embedded Human-in-the-Loop) \cite{Persello2015}.   
Our motivation for this case study is to provide a proof-of-concept for setting up a model for automatic pre-selection of images from large unlabeled datasets that may be used for training ConvNets to learn the visual clues that are indicative of the quality of real estates. This work is therefore intended to serve as the basis for a more extensive follow-up study. Thus, the main incentive is to investigate how the proposed model processes complex intrinsic properties of real estate photographs, as well as which domain-specific labels are generalized well by the classifiers.

Real estate images have different resolutions or were taken under different lighting conditions with varying distances and angles to the object. An additional challenge is that there are only a limited number of relevant labels, and it is a priori unclear which classes can even be captured from the images. The data contains noise, samples that cannot be attributed to a specific property characteristic, as well as redundant information because real estate developers in local markets often work with multiple agencies for advertising and sales.

\section{APPROACH}
We make the naive assumption that empirical error in the decision boundary can be minimized by exploiting the generalization capability of multiple ConvNets, provided that a large amount of training data is available. In this regard, we propose a SSL procedure as follows.

\subsection{Iterative training}
We use annotated data to iteratively fine-tune VGG16 \cite{Simonyan2015} and ResNet101v2 \cite{He2016} (both pre-trained on the large ImageNet dataset), starting from the initial training dataset $S_i$. That is, after each complete iteration, we infer labels in the unlabeled dataset $\emph{\hausaD}_\emph{\textupsilon}$ with  fine-tuned networks $N_1$ and $N_2$ and enrich training datasets $S_1$ and $S_2$ (one set per network) with new instances. Thereby, we select randomly, at a lower threshold of 100\% accuracy, 5 predictions per class and network and add them as new instances to the prior training sets. This process is performed sequentially until we obtain training sets $S_1$ and $S_2$ with 5000 instances each. The selection of  5 matches per class is deliberate to reduce the target risk due to the learner's prior knowledge \cite{Poura2015}. The determination of false predictions in the $S_1$ and $S_2$ is carried out within the definition of experiment baselines (see IV-C).

\subsection{Ensemble decision}
We build a dataset $S_{tr}$ consisting solely of instances of $S_1$ and $S_2$ that are  predicted in concordance by both networks. The inference of the SSL model is then evaluated by fine-tuning a VGG16 with $S_{tr}$ and testing it with an independent dataset $T_1$.

\section{EXPERIMENTAL SETUP}

\subsection{Data}
The preprocessing of the data initially involves duplicate removal by image-wise assignment of unique hash values and calculating difference using Hamming distance. After this step, our experimental data set $\emph{\hausaD}_\emph{\textupsilon}$ eventually comprises 47k images. However, some redundant information remains, as agencies often add their logos when editing photos or post-processing the image for marketing purposes.

\begin{figure}[thpb]
  \centering
  % \framebox{\parbox{3in}{aaaa }}
  \includegraphics[width=0.39\textwidth,height=4cm]{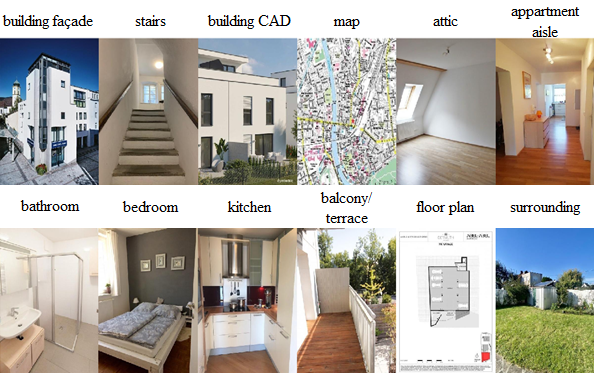}
  \caption{Experimental selection of real estate classes, Image source: \cite{justimmo}}
  \label{figurelabel}
\end{figure}

For our study, we use a manually pre-selected ground truth set $\emph{\hausaD}_\emph{\textiota}$ with 12 meaningful classes from the perspective of real estate valuation. Figure \ref{figurelabel} shows the experimental class selection.
This set is then partitioned into training $S_i$, validation $V_1$ and test $T_1$ datasets with a ratio of 1473-375-240 instances and 12 balanced classes per set. 
We control our experiment by setting multiple baselines (see \ref{eval}) with training sets $S_i$, $S_3$  and $S_4$ (see Table \ref{table:figurelabel2}).
$S_3$ is a manually selected subset of $S_1$ where only correctly predicted labels are kept. $S_4$ is defined like $S_3$ with the exception that the incorrectly predicted labels are not excluded but manually added to the images with correctly predicted labels from $S_1$.

\subsection{Setup \& Training}
For the training we utilize extensive data augmentation including centering, rescaling and shifting. 
Training parameters for both nets are learning rate of 0.001, decay of 0.001, momentum of 0.9 and a batch size of 40 for $N_1$ resp. 100 for $N_2$ . All nets were trained with cross-entropy loss and adamax optimizer \cite{Kingma2015}. A full SSL iteration was initially set to 200 epochs and successively reduced: $I_1=200, I_2=200//2, I_3=200//3, I_4=200//4,...,I_n=200//4$. Since we observed higher loss/accuracy variability in the earlier and later training phases, a larger number of epochs was deliberately chosen. Thus, we do not apply early stopping for regularization but select the training stage with the best performance.

\subsection{Evaluation} \label{eval}
We aim at answering following research questions: (1) are the individual classes sufficiently discriminative to achieve an acceptable generalization of the classifier? and (2) can the proposed experimental SSL approach achieve a comparable result to the established baselines? 
To measure the performance of the model, we set up multiple baselines whose performance was evaluated with the test set $T_1$.  The lower baseline is defined as the performance of a fine-tuned VGG16 trained on initial training set $S_i$. The mid baseline is specified through the performance of a fine-tuned VGG16 trained on $S_3$. Finally, we define an upper baseline as the performance of a fine-tuned VGG16 trained on $S_4$.

\section{RESULTS}
\begin{figure}[thpb]
  \centering
  % \framebox{\parbox{3in}{aaaa }}
  \includegraphics[width=0.32\textwidth,height=5.4cm]{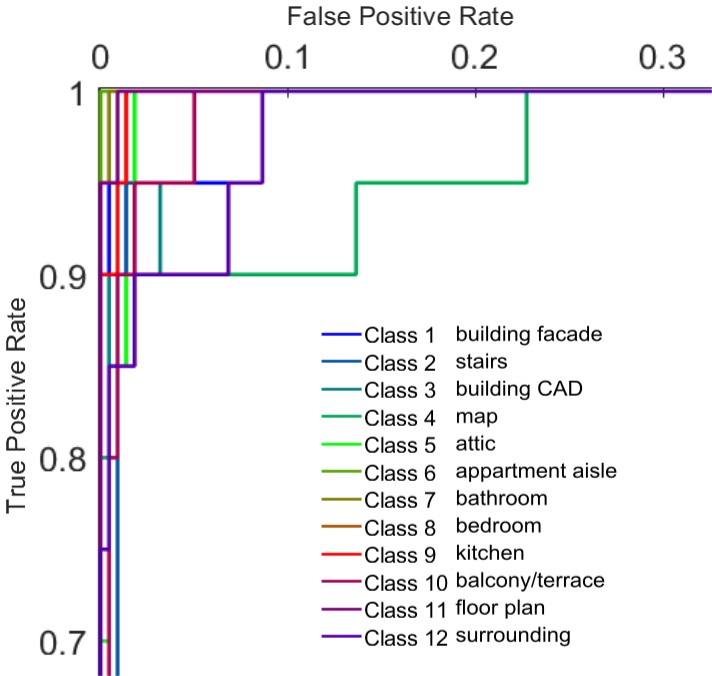}
  \caption{ROC of individual classes.}
  \label{figurelabel3}
\end{figure}

In the Figure \ref{figurelabel3} showing Receiver Operating Characteristic (ROC) for each predicted class, a larger deviation is noticable for class 4 (map), followed by class 12 (surrounding) and class 10 (balcony/terrace). These are basically classes that do not represent interior spaces. An expected confusion can be seen between class 1 (building facade) and class 3 (building CAD). On the other hand, all classes with interiors were particularly well recognized by the classifier, indicating their discriminative visual content. However, false-positive test results point to a minor misinterpretation for classes attic and staircase.

Table \ref{table:figurelabel2} shows that the SSL model slightly underperforms lower and middle baseline, but the performance is almost consistent with the upper baseline. 

This is attributed to the larger proportion of false positives for classes stairs, building facade and building CAD in $S_{tr}$ (compared to $S_{1}$ and $S_{2}$) and thus the inconsistent class balance during the training. Notably, the overall class balance in $S_{tr}$ (intentionally not supervised) expressed by coefficient of variation CV (18\%) is smaller than CV for $S_3$ (30.2 \%) and $S_4$ (41.7 \%).

With this study, we have gained first insights into the challenging task of enriching metadata from real estate images. We intend to build on the results of the presented approach in a  more comprehensive follow-up study to gain further valuable evidence.

\begin{table}[thpb]
  \centering % to have it centered
  \caption{Comparison of classification accuracy (in \%) for SSL model and base models.}\label{tab:somelabel}
  \includegraphics[width=0.49\textwidth, height=2.9cm]{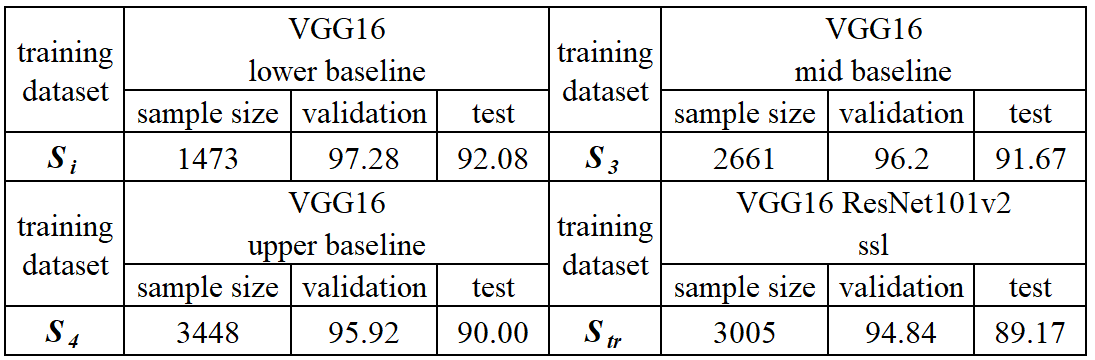}
    \label{table:figurelabel2}
\end{table}

% \section{CONCLUSIONS}
% This command serves to balance the column lengths on the last page of the 
% document manually. It shortens  the textheight of the last page by a   
% suitable amount. This command does not take effect until the  next page so it 
% should come on the page before the  last. Make sure that you do not shorten 
% the textheight too much. 
\addtolength{\textheight}{-12cm}
%%%%%%%%%%%%%%%%%%%%%%%%%%%%%%%%%%%%%%%%%%%%%%%%%%%%%%%%%%%%%%%%%%%%%%%%%%%%%%%%
{\small
\bibliographystyle{IEEEtranS}
\bibliography{main}
}
\end{document}